\documentclass{article}

 \usepackage[main, final]{neurips_2026}

\PassOptionsToPackage{numbers, compress}{natbib}
\usepackage[utf8]{inputenc} 
\usepackage[T1]{fontenc}    
\usepackage{hyperref}       
\usepackage{url}            
\usepackage{booktabs}       
\usepackage{amsfonts}       
\usepackage{nicefrac}       
\usepackage{microtype}      
\usepackage{xcolor}         

\usepackage{graphicx}
\usepackage{booktabs}
\usepackage{caption}
\usepackage{multirow}
\usepackage{multicol}
\usepackage{subfloat}
\usepackage{amsmath}
\usepackage{amssymb}
\usepackage{booktabs}
\usepackage{wrapfig}
\usepackage{colortbl}
\usepackage{makecell}

\usepackage{pifont}
\newcommand{\cmark}{\ding{51}}
\newcommand{\xmark}{\ding{55}}

\title{Pose-Free Feed-Forward 3D Inpainting via Learnable Mask Attention and Support Token Refinement}

\author{%
  \textbf{Jingyi Pan}$^{1}$ \quad
  \textbf{Dan Xu}$^{2}$\thanks{Corresponding author.} \quad
  \textbf{Qiong Luo}$^{1,2}$\footnotemark[1] \\[0.4em]
  $^{1}$The Hong Kong University of Science and Technology (Guangzhou) \\
  $^{2}$The Hong Kong University of Science and Technology \\[0.3em]
  \texttt{jpan305@connect.hkust-gz.edu.cn} \\
  \texttt{\{danxu,luo\}@cse.ust.hk}
}

\begin{document}

\maketitle

\begin{abstract}
    3D scene inpainting aims to recover missing or occluded regions in edited 3D scenes, while ensuring geometric and textural consistency. Existing approaches, however, typically require accurately calibrated camera poses, which restricts their applicability in casual, in-the-wild scenarios and introduces additional preprocessing overhead. To overcome this limitation, we present \textbf{FreeInpaint}, a novel feed-forward framework that generates complete and 3D-consistent scenes directly from unposed multi-view images with masked regions. 
    At its core, FreeInpaint extends a 3D foundation model to propagate masked regions from a reference view to other unposed views, bridging 3D reconstruction and scene inpainting while preserving the model's native ability to recover camera poses and scene geometry.
    Our method addresses two key challenges in adapting feed-forward 3D foundation models to masked inputs. First, masked regions can corrupt cross-view correspondence reasoning, degrading pose estimation and geometry recovery. To address this, we introduce a Learnable Mask Attention mechanism that preserves the spatial anchoring of reliable observations while allowing masked regions to progressively absorb useful context in deeper layers. Second, under severe occlusions, a single forward pass often lacks sufficient appearance evidence for high-fidelity completion. Therefore, we propose a Support Token Refinement strategy, which injects diffusion-generated support evidence as confidence-weighted auxiliary tokens to refine under-observed regions while preserving the original spatial anchor. Extensive experiments across diverse datasets demonstrate that FreeInpaint achieves superior inpainting quality, eliminating the reliance on pre-computed camera poses while keeping a fast inference speed. The project page is \url{https://rorisis.github.io/FreeInpaint/}.
\end{abstract}

\section{Introduction}
\label{sec:intro}
The increasing demand for 3D content creation in virtual reality, augmented reality, and interactive gaming has stimulated growing interest in 3D scene editing. Among various editing tasks, 3D scene inpainting aims to remove unwanted objects or complete missing regions while preserving geometric and textural consistency across multiple views. Compared with 2D inpainting, which operates on a single image, 3D inpainting must maintain multi-view consistency in both appearance and geometry, making it substantially more challenging.

Despite recent progress, existing 3D scene inpainting approaches~\cite{mirzaei2023spin,mirzaei2023reference, wang2024gscream, liu2024infusion, lin2024taming,chen2024mvip,pan2025diga3d, wu2025aurafusion360, huang20253d,you2025instainpaint} suffer from a challenge: they rely on precisely calibrated camera poses. Typically, obtaining these poses requires running off-the-shelf Structure-from-Motion (SfM)~\cite{schonberger2016structure}, such as COLMAP, as a pre-processing step. However, this multi-stage approach is inherently time-consuming, restricting the efficiency and interactivity expected in user-facing inpainting applications.
Furthermore, the reliance on SfM introduces a practical bottleneck for casual, in-the-wild captures. SfM pipelines are brittle and frequently fail when presented with common real-world conditions, such as sparse view collections, large untextured regions, or masked observations. If this initial pose estimation step fails to register cameras or yields inaccurate calibrations, the downstream 3D inpainting methods are fundamentally compromised, inevitably producing visual artifacts or distorted geometry.

\begin{figure}[t]
    \centering
    \includegraphics[width=\linewidth]{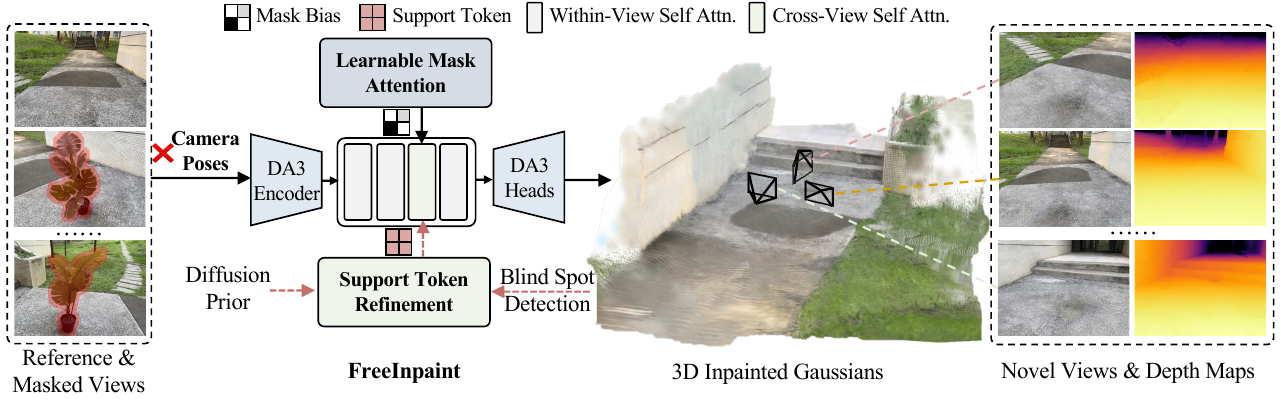}
    \vspace{-10pt}
    \caption{\small \textbf{FreeInpaint} takes a single reference image and multiple unposed masked source images as inputs. It jointly estimates the camera parameters and reconstructs a high-fidelity, multi-view consistent 3D Gaussian field.}
    \label{fig:coverfigure}
    \vspace{-18pt}
\end{figure}
To address this limitation, we present \textbf{FreeInpaint}, a novel feed-forward, reference-based framework that generates complete, 3D-consistent scenes directly from unposed image collections containing user-defined masked regions, as shown in Fig.~\ref{fig:coverfigure}. 
At its core, our method builds upon the architecture of a 3D foundation model~\cite{lin2026depth}, and adapts it for inpainting by conditioning on masked, unposed views while preserving the backbone's strong geometric priors.
 
However, adapting a reconstruction-oriented foundation model for inpainting introduces two primary technical challenges.
First, masked regions can disrupt the multi-view correspondences essential for accurate pose and geometry estimation. If processed without constraints, these masked areas inevitably corrupt the spatial anchoring of the surrounding valid context.
Second, a single feed-forward pass is often insufficient for handling severe occlusions where masked regions lack visual cues from the reference view, i.e., blind spots. In such cases, forcing the network to inpaint highly ambiguous regions in one step typically results in blurry or degraded textures.

To tackle these challenges, FreeInpaint introduces two complementary components.
First, to prevent the corruption of pose estimation and geometry recovery under masked inputs, we introduce a \textbf{Learnable Mask Attention} mechanism. In the early layers of the network, masked regions carry little meaningful information; thus, valid tokens should not attend to them during the self-attention process. However, as the network deepens, these masked regions gradually aggregate contextual information from the reference view and surrounding tokens, eventually becoming useful for global reasoning. To model this dynamic transition, we apply a learnable mask bias term to the attention logits associated with masked tokens within each self-attention block, allowing the network to adaptively regulate how masked regions participate in cross-view reasoning at different depths.
Second, we propose a \textbf{Support Token Refinement} strategy to improve blind-spot completion at inference time. When a single feed-forward pass remains under-confident in severely occluded regions, we inject generated support evidence as confidence-weighted support tokens. This allows the model to refine under-observed regions with complementary appearance cues while preserving the original reference-driven structure and multi-view consistency. Extensive experiments on SPIn-NeRF~\cite{mirzaei2023spin}, 360-USID~\cite{wu2025aurafusion360}, and LLFF~\cite{mildenhall2019llff} datasets demonstrate that FreeInpaint achieves superior performance in both forward-facing and 360-degree scenes, all without requiring camera poses.

In summary, our key contributions are:
(\textbf{i}) We propose FreeInpaint, a feed-forward framework capable of performing 3D scene inpainting directly from unposed, masked image collections.
(\textbf{ii}) We introduce a Learnable Mask Attention mechanism that prevents masked regions from corrupting the foundation model’s native pose and geometry estimation.
(\textbf{iii}) We introduce a Support Token Refinement strategy that injects diffusion-generated support evidence as auxiliary tokens for blind-spot inpainting.
(\textbf{iv}) We demonstrate the effectiveness of our method through extensive experiments on diverse datasets, highlighting its ability to eliminate the reliance on pre-computed camera poses.

\section{Related Work}
\label{sec:related}
\noindent\textbf{3D Scene Inpainting.}
The rapid progress in neural scene representations~\cite{mildenhall2021nerf, wang2021neus, kerbl20233d} has catalyzed significant advancements in 3D scene inpainting~\cite{mirzaei2023spin, chen2024mvip, weber2024nerfiller, liu2024infusion, cao2024mvinpainter, huang20253d}. The primary aim of 3D inpainting is to fill missing regions within a 3D scene, which includes tasks such as object removal and generating realistic textures and geometries to complete affected areas. Existing approaches can generally be classified into non-reference-based and reference-based methods. Non-reference-based techniques often leverage 2D foundation models like CLIP~\cite{radford2021learning} or DINO~\cite{caron2021emerging} to capture 3D semantics for localized editing~\cite{yang2021self, qin2024langsplat, zhou2024feature}, or employ diffusion priors and Score Distillation Sampling (SDS) to iteratively inpaint missing content~\cite{mirzaei2023spin, chen2024mvip, xiao2025localized}. In contrast, reference-based methods~\cite{mirzaei2023reference, wang2024gscream, lu2024view, liu2024infusion, shi2025imfine, wu2025aurafusion360, pan2025diga3d, you2025instainpaint} utilize a provided reference image to propagate the desired texture and geometry across all viewpoints. However, most existing methods rely on computationally intensive per-scene optimization and accurate camera poses, typically obtained via Structure-from-Motion (SfM)~\cite{schonberger2016structure}, limiting their generalization to casual, sparse inputs. Unlike these methods, we propose FreeInpaint, a feed-forward and pose-free framework for 3D scene inpainting.

\noindent\textbf{3D Feed-Forward Models.}
Recent years have witnessed a paradigm shift toward feed-forward 3D reconstruction models, which directly map 2D images to 3D representations without relying on per-scene optimization~\cite{yin2023metric3d, bochkovskii2024depth}. A pioneering work in this domain is DUSt3R~\cite{wang2024dust3r}, which introduces a transformer-based architecture to process unposed image pairs and predict dense 3D pointmaps. Subsequent works~\cite{bhat2023zoedepth, leroy2024grounding, hong2023lrm, zhang2024gs} have explored broader and more complex scenarios. Furthermore, VGGT~\cite{wang2025vggt} presents a unified architecture utilizing alternating intra-frame and inter-frame attention mechanisms across the entire sequence, enabling the joint prediction of camera poses, depth maps, and point correspondences. Following this trajectory, recent foundation models~\cite{keetha2025mapanything, lin2026depth}, particularly Depth Anything 3 (DA3), employ a GS-DPT head to further extend the capability of feed-forward 3D reconstruction to high-fidelity novel view synthesis. However, these models inherently assume a fully observed scene and lack the generative capacity required for dense inpainting in severely occluded areas. In our work, we exploit the robust geometric priors of these feed-forward foundation models and adapt them to pose-free 3D scene inpainting.

\section{Method}
\label{sec:method}
\begin{figure}[t]
    \centering
    \includegraphics[width=\linewidth]{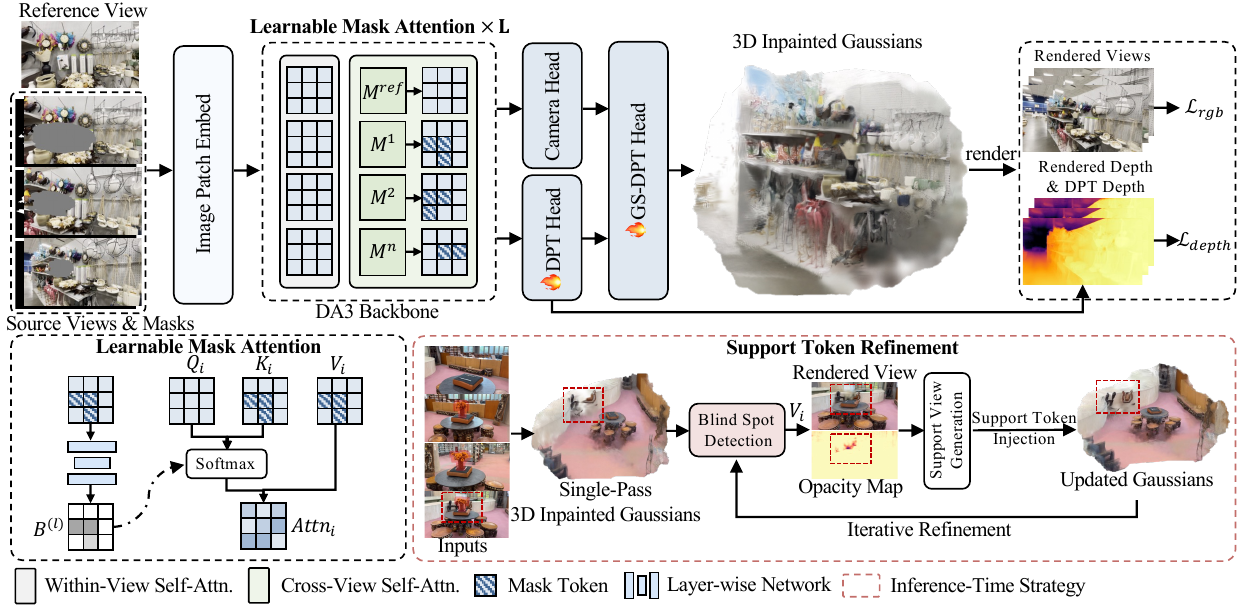}
    \vspace{-12pt}
    \caption{\small \textbf{The pipeline of our FreeInpaint.} Our method adapts DA3 for pose-free 3D inpainting by introducing learnable mask attention, which prevents masked regions from corrupting both pose and geometry estimation (Sec.~\ref{sec:mask_attn}). Additionally, we propose an inference-time support token refinement strategy to progressively inpaint content in occluded blind spots lacking visual cues (Sec.~\ref{sec:iter_refine}), while fine-tuning the DPT heads to adapt the model for inpainting tasks (Sec.~\ref{sec:opt}).}
    \vspace{-15pt}
    \label{fig:framework}
\end{figure}

As illustrated in Fig.~\ref{fig:framework}, given a single reference image $I_\text{ref}$ and a set of source images $\{I_\text{src}^i\}_{i=1}^N$ with corresponding binary masks $\{M^i\}_{i=1}^N$ indicating regions to be inpainted, our goal is to reconstruct a comprehensive 3D scene representation $\mathcal{G}$ (parameterized as 3D Gaussians). Notably, we assume the camera poses for all inputs are unknown. The model must implicitly estimate relative poses while simultaneously inpainting geometry and texture within the masked regions of $I_\text{src}$ to ensure consistency with $I_\text{ref}$, enabling high-fidelity novel view synthesis from arbitrary viewpoints.

This section is organized as follows: Sec.~\ref{sec:preliminary} introduces the preliminaries of Depth Anything 3~\cite{lin2026depth} and 3D Gaussian Splatting~\cite{kerbl20233d}. Sec.~\ref{sec:mask_attn} details our learnable mask attention mechanism, which prevents masked regions from corrupting pose and geometry estimation. Sec.~\ref{sec:iter_refine} describes the support token refinement for handling occlusions. Finally, Sec.~\ref{sec:opt} outlines the training objectives and optimization strategy.

\subsection{Preliminary}\label{sec:preliminary}
\noindent\textbf{Depth Anything 3.}
Our framework builds upon Depth Anything 3 (DA3)~\cite{lin2026depth}, a recent visual geometry foundation model designed to recover consistent 3D spatial representations from an arbitrary number of images. 
Given an input image set $\mathcal{I}=\{I_{i}\}_{i=1}^{N}$, DA3 employs a Vision Transformer backbone (\textit{i.e.}, DINOv2~\cite{oquab2023dinov2}) where each view is prepended with a learnable camera placeholder token to handle unposed scenarios. These tokens are processed via alternative attention blocks that switch between within-view and cross-view self-attention to aggregate multi-view consistency. For output, DA3 utilizes specialized heads to decode these refined features. A Dual-DPT head predicts dense depth and ray maps, while a lightweight camera head regresses explicit camera intrinsics and extrinsics from the refined camera tokens. In addition, a GS-DPT head predicts pixel-aligned 3D Gaussian parameters (i.e., opacity, rotation, scale and colors). This architecture enables robust feed-forward reconstruction and novel view synthesis from unposed images.

\noindent\textbf{3D Gaussian Splatting.}
3D Gaussian Splatting (3DGS)~\cite{kerbl20233d} represents a scene as a set of explicit anisotropic Gaussians, each parameterized by a center $\mu$, opacity $\alpha$, color $c$, rotation $R$, and scale $S$. Unlike implicit radiance fields~\cite{mildenhall2021nerf, wang2021neus}, 3DGS renders novel views by projecting Gaussians onto the image plane and alpha-compositing the projected primitives.

\subsection{Learnable Mask Attention}\label{sec:mask_attn}
A central challenge in pose-free 3D scene inpainting is that, although current 3D foundation models excel at jointly predicting camera poses and 3D geometry, they are highly vulnerable to masked regions. These regions contain no valid information and severely degrade the accuracy of pose and geometry estimation if processed naively. 

A straightforward strategy is to apply a hard binary attention mask that permanently blocks masked regions. However, such a formulation is overly restrictive for inpainting. As information propagates through the network, some masked tokens may gradually become informative due to context aggregation from neighboring valid regions and other views. Permanently suppressing them prevents the model from exploiting this evolving evidence. Previous works~\cite{barrios2024multi, li2022mat, liu2021swin} have explored dynamic or learnable masking mechanisms to modulate information propagation in image inpainting or more general vision tasks. However, MAT~\cite{li2022mat} relies on heuristic mask-update rules, MLLAM~\cite{barrios2024multi} predicts independent token weights at each layer, and Swin Transformer~\cite{liu2021swin} employs relative-position biases and structural attention masks for shifted-window computation. 

To address these, we design a learnable mask attention mechanism to encourage the model to preserve the spatial anchoring of reliable observations while limiting the corruption introduced by masked tokens. Specifically, it introduces a per-token, key-side attention bias initialized from explicit input occlusion masks and progressively updated across layers through learnable residuals. Let $M^i$ denote the binary mask of source view $i$. We first downsample each mask to the patch resolution of the DA3 transformer and flatten it into a token-level vector $\tilde{M}^i$. Rather than treating $\tilde{M}^i$ as a fixed masking rule, we use it to initialize a sequence of learnable soft masks that evolve across the cross-view self-attention layers. 

For each cross-view attention layer $l$, we derive a token-level mask bias $B^{(l)}$ from the corresponding source tokens and binary masks. This bias is initialized from the binary mask and then progressively refined through a lightweight layer-wise network $g_l$ conditioned on the current source tokens: 
\begin{align}
P^{(0)} &= \operatorname{logit}
\left(\epsilon + (1-2\epsilon)\tilde{M}\right), \\
P^{(l)} &= P^{(l-1)} + \alpha g_l\left(F^{(l)}\right),\\
B^{(l)} &= -\beta\sigma\left(P^{(l)}\right),
\end{align}
where $P^{(l)}$ denotes the token-level suppression state at layer $l$, $F^{(l)}$ denotes the current source tokens, $\sigma(\cdot)$ denotes the sigmoid function, $\operatorname{logit}(p)=\log\frac{p}{1-p}$, and $\epsilon$ avoids infinite logits for binary mask values. Furthermore, given the query, key, and value $Q$, $K$, and $V$, the cross-view attention is computed as:
\begin{equation}
    \mathrm{Attention}(Q,K,V)=\mathrm{Softmax}\left(\frac{QK^\top}{\sqrt{d}} + B^{(l)}\right)V,
\end{equation}
where $d$ is the key dimension. In this way, the model learns not only \emph{where} masked regions should be suppressed, but also \emph{how strongly} they should participate in cross-view reasoning at different depths.



\subsection{Support Token Refinement}\label{sec:iter_refine}
\begin{figure}[t]
    \centering
    \includegraphics[width=\linewidth]{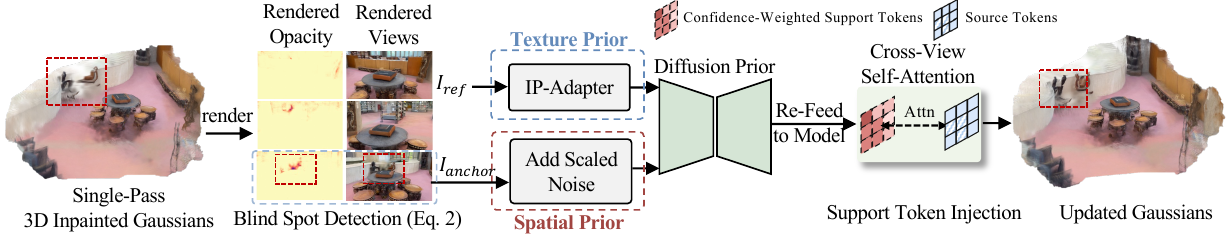}
    \vspace{-10pt}
    \caption{\small \textbf{Support Token Refinement.} We dynamically detect unobserved blind spots and designate them as anchor views. A diffusion prior is then employed to rapidly inpaint the missing content, while spatial and texture priors ensure multi-view consistency during the generation process.}
    \label{fig:iterative}
    \vspace{-10pt}
\end{figure}

While the learnable mask attention mechanism effectively mitigates invalid context propagation, a single feed-forward pass remains insufficient for handling severe occlusions where source views lack visual cues from the reference image. We term these unobserved regions as blind spots. To address this challenge, we introduce a support token refinement strategy at inference time, as shown in Fig.~\ref{fig:iterative}. The key idea is to identify under-observed regions from the current 3D prediction, generate auxiliary support evidence for those regions, and re-inject this evidence into the model as confidence-weighted support tokens.

\noindent\textbf{Blind Spot Detection.}
After an initial feed-forward pass, the predicted 3D Gaussian field is rendered back to the masked source views. For each masked source view $i$, we compute a visibility score over the masked region:
\begin{equation}
V_i = \frac{\sum \left( \alpha^i_{\mathrm{render}} \odot M^i \right)}{\sum M^i + \epsilon},
\end{equation}
where $\alpha^i_{\mathrm{render}}$ denotes the rendered opacity map and $M^i$ is the corresponding binary mask. Low opacity inside the masked region indicates that the current 3D reconstruction lacks sufficient support in that area. We identify the view with the lowest visibility score as the anchor view for refinement.

\noindent\textbf{Support View Generation.}
For the selected anchor view, we introduce a diffusion prior to generate a complementary support view conditioned on the current scene context. To preserve consistency with the existing reconstruction, we apply Stochastic Differential Editing (SDEdit) \cite{meng2021sdedit} using the previously rendered image $I_\text{render}$ as the canvas. By adding noise proportional to a strength factor, we retain the underlying geometry, thereby introducing spatial prior. Simultaneously, we condition the inpainting on the original reference image $I_\text{ref}$ via an IP-Adapter~\cite{ye2023ip} to further inject the textural prior. The resulting support view generated by a diffusion model with both spatial and textural priors thus supplies complementary appearance cues for the blind spot while remaining consistent with the original reference image.

\noindent\textbf{Support Token Injection.}
The generated support view is encoded into auxiliary tokens, which are then injected into the model in a confidence-weighted manner to refine under-observed regions. To determine the reliability of these support tokens, we compute a confidence weight map. Instead of trusting the generated support view uniformly over the masked region, we prioritize true blind spots by combining the anchor mask with the rendered opacity:
\begin{equation}
    C^{\mathrm{sup}} = \mathrm{Norm}\!\left(\mathcal{G}_{\sigma}\!\left(M^{a}\odot(1-\alpha^{a}_{\mathrm{render}})^{\gamma}\right)\right),
\end{equation}
where $M^{a}$ is the anchor mask, $\alpha^{a}_{\mathrm{render}}$ is the rendered opacity map of the anchor view, $\mathcal{G}_{\sigma}$ denotes Gaussian smoothing, and $\mathrm{Norm}(\cdot)$ rescales the response to $[0,1]$. In this way, regions already explained by the current reconstruction receive low support confidence, while low-opacity blind spots receive high confidence. Finally, we resize 
 to the token resolution and use it to modulate the attention given to the support tokens, implementing the confidence-weighted injection. Consequently, high-confidence support tokens from blind spots are effectively attended to, whereas low-confidence ones remain suppressed, allowing the model to refine blind spots without disturbing the original reference-driven structure.

\subsection{Optimization}\label{sec:opt}
To effectively adapt the 3D foundation model from a reconstruction-based model to an inpainting-capable framework, we detail our fine-tuning strategy and training objectives.

\noindent\textbf{Mask Generation.}
To simulate realistic occlusion patterns during training, we employ a hybrid masking strategy. For each training clip, we designate one frame as the reference view $I_\text{ref}$, while others serve as source views  $I_\text{src}$ or novel views $I_{novel}$. We generate masks $M$ for $I_\text{src}$ using two methods: a) \textit{Random Box and Brush Masks}, which simulate arbitrary occlusions ranging from simple artifacts to large-scale missing regions; and b) \textit{Geometric Masks}, where 2D masks are projected from the reference view to source views using the metric depth estimated by the original pre-trained DA3~\cite{lin2026depth}. This simulates physically consistent 3D occlusions, compelling the model to learn geometry-aware inpainting rather than simple 2D texture infilling.

\noindent\textbf{Depth-Guided Self-Distillation.}
A major challenge in feed-forward 3D inpainting is recovering accurate geometry within occluded regions where photometric constraints are entirely absent. Without explicit geometric guidance, models tend to hallucinate textures onto incorrect depth planes. To mitigate this, we formulate the geometric reconstruction as a self-distillation process, leveraging the strong spatial priors embedded in the original foundation model.
During training, we employ a frozen copy of the pre-trained DA3 as a teacher network. As the model processes a scene, we pass the unmasked source and novel views through this frozen teacher to generate high-quality depth maps, denoted as $D_\text{prior}$, on the fly. These serve as pseudo ground-truth. Simultaneously, our trainable model processes the masked images to predict the inpainted depth, $D_\text{pred}$. We enforce geometric consistency between the student's prediction and the teacher's prior via a masked distillation loss:
\begin{equation}
    \mathcal{L}_\text{depth} = \| {D}_\text{pred} \odot M - {D}_\text{prior} \odot M \|_1,
\end{equation}
where ${D}$ denotes the normalized depth, and $M$ is the binary mask.

\begin{table*}[t]
    \centering
    \setlength{\tabcolsep}{0.03\linewidth}
    \caption{\small Quantitative comparisons with per-scene optimization methods on SPIn-NeRF~\cite{mirzaei2023spin}, 360-USID~\cite{wu2025aurafusion360}, and LLFF~\cite{mildenhall2019llff} datasets. Inference time is measured on a single A6000 GPU. We report the cumulative time for multi-stage pipelines, while our method requires only $\sim$0.4s for a single forward pass. \textsuperscript{*}For scenes with severe occlusions, a dynamic Support Token Refinement is triggered, slightly increasing the time to $\sim$1.8s (see Sec. 4.2), which remains orders of magnitude faster than optimization baselines.}
    \vspace{-5pt}
    \resizebox{\linewidth}{!}{
    \begin{tabular}{l|cc|cc|cc|c}
    \toprule[1.2pt]
    \multirow{2}{*}{Methods}& \multicolumn{2}{c|}{SPIn-NeRF~\cite{mirzaei2023spin}} & \multicolumn{2}{c|}{360-USID~\cite{wu2025aurafusion360}} & \multicolumn{2}{c|}{LLFF~\cite{mildenhall2019llff}} & \multirow{2}{*}{Time$\downarrow$}\\
         &PSNR $\uparrow$ & LPIPS$\downarrow$ & PSNR$\uparrow$ & LPIPS$\downarrow$ & C-KID$\downarrow$ & C-FID $\downarrow$\\ \midrule
         SPIn-NeRF~\cite{mirzaei2023spin} & 16.32 & 0.4122 & 17.59 & 0.4126 & 0.5978&388.29 & $\sim$ 5h\\
         NeRFiller~\cite{weber2024nerfiller} & \underline{16.86} & 0.4183& 15.03&0.5036 & \underline{0.5835}&\underline{360.02}&$\sim$ 30m\\
         InFusion~\cite{liu2024infusion} &13.99&0.5216& 15.62 & 0.3716 &0.6652&428.18&$\sim$ 30m\\
         GScream~\cite{wang2024gscream} & 16.85 & \underline{0.3644} & 16.29 & 0.4415 & 0.6300 &409.16 &$\sim$ 2h\\
         AuraFusion360~\cite{wu2025aurafusion360} &14.51&0.7077& \underline{18.34}& \underline{0.2982}& 0.6378&402.18&$\sim$ 1h\\ \midrule
         \textbf{Ours} & \textbf{17.79}&\textbf{0.2819} & \textbf{18.58}&\textbf{0.2526} &\textbf{0.5613} & \textbf{343.97} & $\sim$ \textbf{0.4s\textsuperscript{*}}\\
    \bottomrule[1.2pt]
    \end{tabular}}
    \label{tab:q1}
    \vspace{-8pt}
\end{table*}

\noindent\textbf{Training Objectives.}
The overall training objective is formulated as a combination of a photometric loss and a multi-view depth consistency loss:
\begin{equation}
    \mathcal{L}_\text{total} = \mathcal{L}_\text{rgb} + \lambda_\text{depth}\mathcal{L}_\text{depth},
\end{equation}
where $\lambda_\text{depth}$ is a balancing weight. The photometric loss $\mathcal{L}_\text{rgb}$ is computed using a MSE loss alongside an LPIPS perceptual penalty:
\begin{equation}
    \mathcal{L}_\text{rgb} = \mathcal{L}_\text{mse} + \lambda_\text{lpips}\mathcal{L}_\text{lpips}.
\end{equation}
To prevent the model from trivially overfitting input views, we compute the photometric loss $\mathcal{L}_\text{rgb}$ exclusively on held-out novel views. In contrast, the depth loss $\mathcal{L}_\text{depth}$ is evaluated on both the input views and the novel views to enforce 3D geometric consistency across the entire scene. 

\begin{figure}[t]
    \centering
    \includegraphics[width=\linewidth]{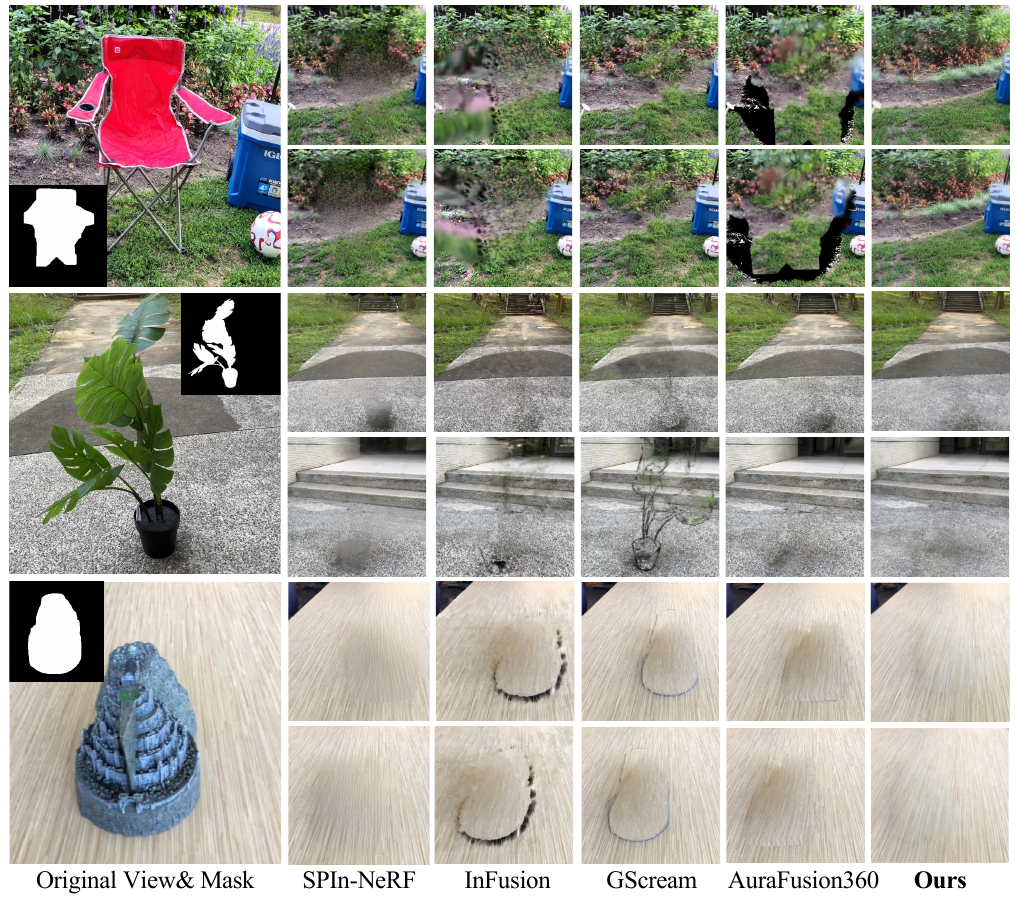}
    \vspace{-15pt}
    \caption{\small Qualitative results with per-scene optimization methods on SPIn-NeRF~\cite{mirzaei2023spin}, 360-USID~\cite{wu2025aurafusion360}, and LLFF~\cite{mildenhall2019llff} datasets. FreeInpaint successfully synthesizes high-fidelity and multi-view consistent novel views without requiring camera poses.}
    \label{fig:removal}
    \vspace{-5pt}
\end{figure}

\begin{table*}[t]
    \centering
    \setlength{\tabcolsep}{0.028\linewidth}
    \caption{\small Quantitative comparisons with feed-forward unposed methods on SPIn-NeRF~\cite{mirzaei2023spin}, 360-USID~\cite{wu2025aurafusion360}, and LLFF~\cite{mildenhall2019llff} datasets.}
    \vspace{-8pt}
    \resizebox{\linewidth}{!}{
    \begin{tabular}{l|ccc|ccc|cc}
    \toprule[1.2pt]
    \multirow{2}{*}{Methods}& \multicolumn{3}{c|}{SPIn-NeRF~\cite{mirzaei2023spin}} & \multicolumn{3}{c|}{360-USID~\cite{wu2025aurafusion360}} & \multicolumn{2}{c}{LLFF~\cite{mildenhall2019llff}}  \\
         &PSNR$\uparrow$ & LPIPS $\downarrow$ & FID$\downarrow$ & PSNR$\uparrow$& LPIPS$\downarrow$ & FID$\downarrow$ & C-KID$\downarrow$ & C-FID $\downarrow$\\ \midrule
         LAMA~\cite{suvorov2022resolution}+DA3 &16.67& 0.4034 & 196.01 & 17.76& 0.3211&186.83 & 0.5629 & 361.75 \\
         MVInpainter~\cite{cao2024mvinpainter}+DA3 &16.79 &0.3733 & 164.22 & 17.24 & 0.3891 & 275.06& 0.5724 & 372.76\\ \midrule
         \textbf{Ours}&\textbf{17.79}&\textbf{0.2819}& \textbf{148.05}& \textbf{18.58}&\textbf{0.2526} & \textbf{199.61}&\textbf{0.5613} & \textbf{343.97}  \\
    \bottomrule[1.2pt]
    \end{tabular}}
    \label{tab:q2}
    \vspace{-10pt}
\end{table*}

\section{Experiments}
\label{sec:exp}
\subsection{Datasets and Evaluation Metrics} 
\noindent\textbf{Datasets.}
We utilize the DL3DV-10K dataset~\cite{ling2024dl3dv} for training, which offers large-scale real-world scenes with diverse camera trajectories. For evaluation, we employ three distinct benchmarks to assess performance across different scenarios: SPIn-NeRF~\cite{mirzaei2023spin}, 360-USID~\cite{wu2025aurafusion360}, and LLFF~\cite{mildenhall2019llff}. 
SPIn-NeRF serves as a standard benchmark for object removal in forward-facing scenes with ground-truth backgrounds. The 360-USID dataset extends this to 360-degree scenes, providing reference images and ground truth to quantitatively evaluate inpainting quality behind removed objects. The LLFF dataset provides diverse real-world scenes with varying view numbers. Additional experiments on other datasets are provided in the Appendix.

\noindent\textbf{Evaluation Metrics.}
Following prior works~\cite{wu2025aurafusion360, shi2025imfine, you2025instainpaint}, we render novel views from the inpainted 3D representations for quantitative assessment. For datasets with available ground truth (SPIn-NeRF and 360-USID), we report PSNR and LPIPS~\cite{zhang2018unreasonable}. For the LLFF dataset, which lacks ground-truth backgrounds for physically removed objects, we employ C-FID~\cite{heusel2017gans} and C-KID~\cite{binkowski2018demystifying} to assess the visual quality and coherence of the inpainted regions. To ensure a focused evaluation, all metrics are computed exclusively within the bounding box defined by the object mask. Furthermore, for text-driven object replacement tasks, we additionally report the CLIP Text-Image Directional Similarity ($\text{CLIP}_\text{dir}$)~\cite{radford2021learning} to evaluate semantic alignment.

\subsection{Implementation Details}
\noindent\textbf{Training.} Our framework is built on Depth Anything 3 (DA3-Giant)~\cite{lin2026depth}. We optimize the model using the DL3DV-10K dataset~\cite{ling2024dl3dv}, sampling 15-frame clips from long video sequences to form each training batch. We adopt a two-stage fine-tuning strategy: we first adapt the DPT and GS-DPT heads to masked scene inpainting, and then further optimize the Learnable Mask Attention modules together with cross-view attention layers. During training, we employ a hybrid masking strategy that mixes random masks and geometric masks at a ratio of 0.4:0.6. Optimization is performed on 8 NVIDIA RTX A6000 GPUs with a per-GPU batch size of 2 and a gradient accumulation step of 2. All input images are resized to a resolution of $280\times504$.

\noindent\textbf{Inference.} During inference, we leverage PowerPaint v1~\cite{zhuang2023task} as a 2D diffusion prior to inpaint the reference image if a clean reference view is initially unavailable. We utilize the Language-based Segment Anything model (Lang-SAM)~\cite{kirillov2023segment} to extract specific object masks. Our support token refinement is dynamically triggered when the rendered visibility score inside the masked region falls below 0.97, with a maximum of 3 total feed-forward passes.

\begin{figure}[t]
    \centering
    \includegraphics[width=\linewidth]{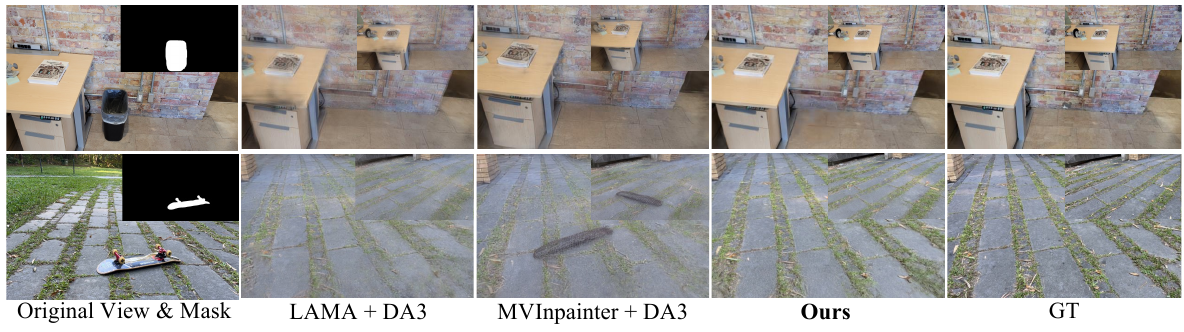}
    \vspace{-14pt}
    \caption{\small Qualitative results on SPIn-NeRF~\cite{mirzaei2023spin} and 360-USID~\cite{wu2025aurafusion360} datasets. }
    \label{fig:removal_2}
    \vspace{-8pt}
\end{figure}

\begin{table*}[t]
\begin{minipage}[t]{0.48\textwidth}
    \centering
    \setlength{\tabcolsep}{0.03\linewidth}
    \caption{\small Quantitative comparisons of the object replacement task.}
    \vspace{-5pt}
    \resizebox{\linewidth}{!}{
    \begin{tabular}{l|c|ccc}
    \toprule[1.2pt]
    Methods& Category&$\text{CLIP}_{\text{dir}}$$\uparrow$& C-KID$\downarrow$ & C-FID $\downarrow$\\ \midrule
         DiGA3D~\cite{pan2025diga3d} & Non-Ref. &0.1928 & 0.2607 &169.08 \\
         NeRFiller~\cite{weber2024nerfiller} &Ref.& 0.1917& 0.2628&198.25\\
         InFusion~\cite{liu2024infusion}& Ref. & 0.1922 & 0.2647& 236.75 \\  \midrule
         \textbf{Ours} & Ref. & \textbf{0.1942}&\textbf{0.2589}&\textbf{153.25}\\
    \bottomrule[1.2pt]
    \end{tabular}}
    \label{tab:q3}
\end{minipage}
\hfill
\begin{minipage}[t]{0.48\textwidth}
    \centering
    \setlength{\tabcolsep}{0.035\linewidth}
    \caption{\small Ablation study on each component on the 360-USID dataset~\cite{wu2025aurafusion360}.}
    \vspace{-5pt}
    \resizebox{\linewidth}{!}{
    \begin{tabular}{l|ccc}
    \toprule[1.2pt]
    Methods& PSNR$\uparrow$ & LPIPS$\downarrow$ & FID $\downarrow$\\ \midrule
         Baseline & 15.58 & 0.7399 & 392.83\\
         + Fine-Tuning (Sec.~\ref{sec:opt}) & 16.96& 0.2854&259.26\\
         + Mask Attn. (Sec.~\ref{sec:mask_attn})& 18.26& 0.2628&209.86\\
         + Token Refine. (Sec.~\ref{sec:iter_refine})& \textbf{18.58}&\textbf{0.2526}&\textbf{199.61}\\
    \bottomrule[1.2pt]
    \end{tabular}}
    \label{tab:ab}
\end{minipage}
\vspace{-10pt}
\end{table*}


\begin{figure}[!t]
    \centering
    \includegraphics[width=\linewidth]{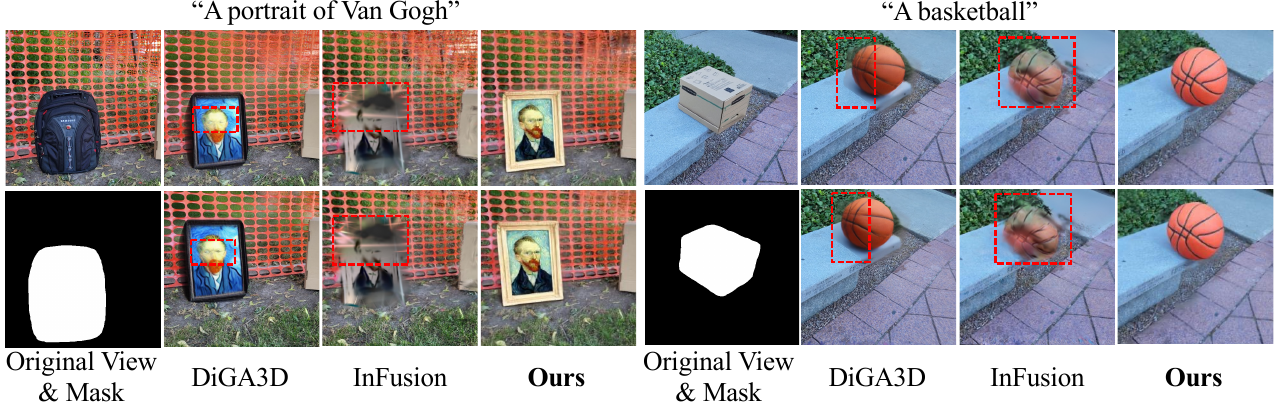}
    \vspace{-15pt}
    \caption{\small Qualitative results of the object replacement task. We compare FreeInpaint with DiGA3D~\cite{pan2025diga3d} and InFusion~\cite{liu2024infusion} methods.}
    \label{fig:replace}
    \vspace{-10pt}
\end{figure}

\subsection{Main Results}
\noindent\textbf{Comparison with Per-Scene Optimization Methods.} Tab.~\ref{tab:q1} presents a quantitative comparison on the SPIn-NeRF~\cite{mirzaei2023spin}, 360-USID~\cite{wu2025aurafusion360}, and LLFF~\cite{mildenhall2019llff} datasets. For our evaluation, FreeInpaint operates on highly sparse and uncalibrated data, utilizing only 4 unposed inputs (1 reference and 3 masked source images) for the SPIn-NeRF and LLFF datasets, and 8 unposed inputs (1 reference and 7 masked source images) for 360-USID dataset. Despite this challenging setting, FreeInpaint consistently outperforms prior methods across most metrics while maintaining rapid inference speeds. Crucially, our feed-forward design enables inference in just 0.4s via a single forward pass, a significant speedup compared to the 30 minutes to 5 hours required by optimization-based approaches. Qualitative results in Fig.~\ref{fig:removal} further validate its ability to generate high-fidelity, consistent renderings under sparse, unposed conditions.

\noindent\textbf{Comparison with Feed-Forward Unposed Methods.} Given the absence of existing feed-forward, pose-free 3D inpainting frameworks, we formulate LAMA~\cite{suvorov2022resolution}+DA3 and MVInpainter~\cite{cao2024mvinpainter}+DA3 for uncalibrated feed-forward comparisons. As reported in Tab.~\ref{tab:q2}, our framework achieves superior performance across all evaluated metrics. As shown in~Fig.\ref{fig:removal_2}, LAMA+DA3 fails due to multi-view inconsistency in 2D inpainting, which corrupts pose estimation and subsequent 3D reconstruction. Even MVInpainter+DA3 exhibits noticeable pose misalignments. These results underscore the effectiveness of our unified approach, which jointly addresses inpainting and pose estimation with sparse and masked inputs, avoiding the error accumulation inherent in cascaded pipelines.

\begin{figure}[!t]
    \centering
    \includegraphics[width=\linewidth]{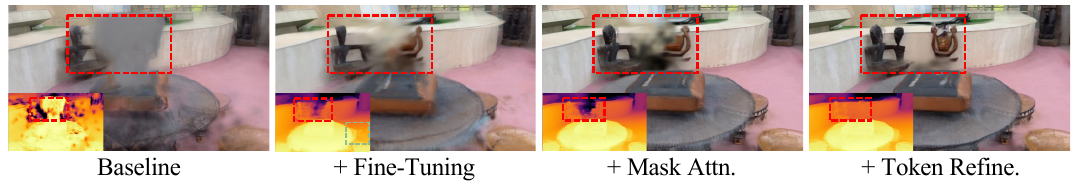}
    \vspace{-18pt}
    \caption{\small \textbf{Qualitative ablation study on the 360-USID dataset.} Fine-tuning (+ Fine-Tuning) alone still yields blurriness. Integrating our learnable mask attention (+ Mask Attn.) significantly sharpens structural details by suppressing invalid features. The support token refinement (+ Token Refine.) further recovers high-fidelity textures in occluded regions.}
    \label{fig:ab}
    \vspace{-12pt}
\end{figure}

\begin{table}[!t]
    \centering
    \small
    \setlength{\tabcolsep}{18pt}
    \caption{\small Effect of Learnable Mask Attention. We compare different mask-aware attention strategies under masked inputs.}
    \label{tab:learnable_mask_attention}
    \begin{tabular}{l|ccccc}
    \toprule
    Mask Attention & PSNR$\uparrow$ & LPIPS$\downarrow$ & ATE$\downarrow$ & RPE$_t$$\downarrow$ & RPE$_r$$\downarrow$ \\
    \midrule
    No mask         & 16.96 & 0.2854 & 0.0462 & 0.0803 & 0.7766 \\
    Hard mask       & 17.97 & 0.2581 & 0.0377 & 0.0764 & 0.7430 \\
    Learnable mask  & \textbf{18.58} & \textbf{0.2526} & \textbf{0.0340} & \textbf{0.0644} & \textbf{0.7069} \\
    \bottomrule
    \end{tabular}
    \vspace{-18pt}
\end{table}

\noindent\textbf{Object Replacement.}
In Tab.~\ref{tab:q3}, FreeInpaint consistently outperforms both the reference-free approach~\cite{pan2025diga3d} and the reference-based methods~\cite{weber2024nerfiller, liu2024infusion}. As illustrated in Fig.~\ref{fig:replace}, previous methods struggle with noticeable artifacts and blurriness within the inpainted regions. In contrast, FreeInpaint synthesizes clean, high-fidelity textures that blend seamlessly with the surrounding 3D context.

\subsection{More Analysis}
\noindent\textbf{Ablation on Main Components.}
We conduct ablation studies on the 360-USID~\cite{wu2025aurafusion360} dataset to validate the effectiveness of each proposed component. As shown in Tab.~\ref{tab:ab} and Fig.~\ref{fig:ab}, directly applying the original pre-trained DA3 as a baseline produces poor results due to the domain gap between standard reconstruction and scene inpainting. Fine-tuning improves basic novel view synthesis, but the inpainted regions remain noticeably blurry, as masked regions can still corrupt cross-view pose and geometry estimation. Introducing learnable mask attention substantially improves all quantitative metrics and visibly reduces blurriness, confirming its ability to preserve spatial anchoring while adaptively incorporating useful context from masked regions. Finally, support token refinement further improves inpainting quality, especially for blind spots lacking sufficient reference-view cues, by progressively injecting high-fidelity auxiliary evidence.

\noindent\textbf{Effect of Learnable Mask Attention.} 
A key contribution of our work is the learnable mask attention mechanism, which not only improves inpainting quality but also enhances pose estimation robustness. To quantify this, we evaluate the camera rotation and translation accuracy using Absolute Trajectory Error (ATE) and Relative Pose Error (RPE) on the 360-USID dataset~\cite{wu2025aurafusion360}. As shown in Tab.~\ref{tab:learnable_mask_attention}, both hard and learnable masking outperform the variant without mask attention, confirming the importance of suppressing invalid masked regions. Moreover, learnable mask attention further improves ATE, RPE$_t$, and RPE$_r$ over fixed hard masking, showing that adaptive mask-aware routing better preserves spatial anchoring while allowing masked regions to absorb useful cross-view context.

\begin{table*}[t]
    \centering
    \begin{minipage}[t]{0.54\textwidth}
        \centering
        \captionof{table}{
        \small View-count scalability on the SPIn-NeRF~\cite{mirzaei2023spin} and 360-USID~\cite{wu2025aurafusion360} datasets.
        ``Total Input Views'' includes one reference view and the remaining masked source views.
        }
        \label{tab:viewscaling}
        
        \resizebox{\linewidth}{!}{%
        \begin{tabular}{c|ccc|ccc}
            \toprule[1.2pt]
            \multirow{2}{*}{\shortstack{Total Input\\Views}}
            & \multicolumn{3}{c|}{SPIn-NeRF}& \multicolumn{3}{c}{360-USID} \\
            & PSNR$\uparrow$& LPIPS$\downarrow$& FID$\downarrow$& PSNR$\uparrow$
            & LPIPS$\downarrow$& FID$\downarrow$ \\
            \midrule
            4& 17.79& 0.2819& 148.05& 17.19& 0.3148& 226.30 \\
            
            8& 18.32& 0.2702& \textbf{143.88}& 18.58& 0.2526& 199.61 \\
            
            16& 18.33& 0.2688& 145.97& 18.66& \textbf{0.2512}& 196.67 \\
            
            32& \textbf{18.47}& \textbf{0.2683}& 146.44& \textbf{18.69} & 0.2519& \textbf{194.32} \\
            \bottomrule[1.2pt]
        \end{tabular}%
        }
    \end{minipage}
    \hfill
    \begin{minipage}[t]{0.44\textwidth}
    \centering
    \captionof{table}{
    \small Sensitivity to the diffusion-generated reference and
    reference view selection on the SPIn-NeRF~\cite{mirzaei2023spin} dataset.
    }
    \label{tab:refsensitivity}

    \resizebox{\linewidth}{!}{%
\begin{tabular}{l|cc}
    \toprule[1.2pt]
    Metric
    & \makecell[c]{Diffusion-generated\\reference}
    & \makecell[c]{Reference view\\selection} \\
    \midrule
    PSNR$\uparrow$
    & $17.68 \pm 0.16$
    & $17.61 \pm 0.24$ \\
    LPIPS$\downarrow$
    & $0.2808 \pm 0.0168$
    & $0.2867 \pm 0.0144$ \\
    FID$\downarrow$
    & $149.85 \pm 7.55$
    & $152.18 \pm 12.76$ \\
    \bottomrule[1.2pt]
\end{tabular}%
}
\end{minipage}
\vspace{-10pt}
\end{table*}

\noindent\textbf{Scalability to More Views.}
To evaluate view-count scalability, we vary the total number of input views from 4 to 32 on SPIn-NeRF~\cite{mirzaei2023spin} and 360-USID~\cite{wu2025aurafusion360} datasets, while keeping all other evaluation settings unchanged. As shown in Tab.~\ref{tab:viewscaling}, FreeInpaint generally benefits from additional input views. The reconstruction quality improves as more unposed source observations are introduced, while the perceptual metrics remain stable with only minor fluctuations. These results show that FreeInpaint is not restricted to the original sparse-view setting and can effectively utilize denser collections of unposed inputs.

\noindent\textbf{Sensitivity to the Reference View.}
We evaluate the sensitivity of FreeInpaint to variations in the reference input on the SPIn-NeRF~\cite{mirzaei2023spin} dataset. Specifically, we consider two sources of variation: 1) diffusion-inpainted references generated with different sampling seeds for a fixed reference viewpoint in each scene, and 2) different choices of the reference viewpoint, which provide varying scene coverage and visible context. We report the mean and standard deviation over ten diffusion seeds and four reference-view choices, respectively. As shown in Tab.~\ref{tab:refsensitivity}, FreeInpaint remains relatively stable under both types of variation, with only moderate changes in reconstruction and perceptual quality. For each setting, we first compute the metrics over the complete test set and then report the mean and standard deviation across the corresponding diffusion seeds or reference-view choices.

\section{Conclusion}
\label{sec:conclusion}
In this paper, we present FreeInpaint, a feed-forward framework for pose-free 3D scene inpainting from uncalibrated and masked image collections. By adapting a 3D foundation model to masked inputs, FreeInpaint enables joint pose estimation, geometry recovery, and scene inpainting without requiring pre-computed camera poses. To prevent masked regions from corrupting spatial anchoring, we introduced a learnable mask attention mechanism that adaptively regulates the participation of masked tokens in cross-view reasoning. To further handle severe occlusions with insufficient reference cues, we proposed an inference-time support token refinement strategy that injects diffusion-generated evidence as auxiliary tokens for progressive inpainting. Extensive experiments demonstrate that FreeInpaint achieves high-quality, multi-view consistent inpainting with fast feed-forward inference, providing an efficient and robust solution for 3D scene inpainting in unposed settings.

\section{Acknowledgments}
The work of Qiong Luo and Jingyi Pan is supported by a startup grant from the Hong Kong University of Science and Technology (Guangzhou). The work of Dan Xu is supported by the Early Career Scheme of the Research Grants Council (RGC) of the Hong Kong SAR under grant No. 26202321, ITF PRP/046/24FX, Science \& Technology Cooperation Program of Shandong under grant No. SDST26EG01, SAIL Research Project, and HKUST-Zeekr Collaborative Research Fund.

\clearpage

{   
    \bibliography{ref}
}

\clearpage


\appendix

\section{Technical Appendices and Supplementary Material}
\subsection{Demo Videos}
To further demonstrate the multi-view consistency and geometric stability of our method, we have included demo video results in our project page.

\subsection{Additional Dataset Details}
In addition to the datasets discussed in the main paper, we incorporate the IMFine dataset~\cite{shi2025imfine} for extended ablation studies and qualitative evaluation in the appendix. IMFine provides complex scenes with challenging view variations, making it particularly well-suited for evaluating the robustness of our method.

\subsection{Additional Implementation Details}
We apply learnable mask attention only to the cross-view attention blocks of DA3. For each cross-view attention block, we instantiate an independent lightweight layer-wise network to predict a learnable mask bias that suppresses invalid masked information from corrupting pose and geometry estimation. The network consists of a $3\times3$ convolution, a depthwise $3\times3$ convolution, and a $1\times1$ projection layer, with GELU activations in between. The final projection layer is zero-initialized such that the learnable mask is initialized as the binary-mask prior and then gradually adapted during training. We set the hidden dimension to 8, the residual scaling factor $\alpha$ to 1.5, and the suppression strength $\beta$ to 10.

\subsection{Additional Details of Feed-Forward Baselines}
For the LaMa+DA3 and MVInpainter+DA3 baselines in Tab.~\ref{tab:q2}, we use the official pretrained checkpoints of LaMa~\cite{suvorov2022resolution} (\texttt{big-lama}) and MVInpainter~\cite{cao2024mvinpainter} (\texttt{MVInpainter-F-512}) without any task-specific fine-tuning. LaMa is applied independently to each masked input view, whereas MVInpainter jointly processes the masked multi-view inputs using its official inference pipeline. The resulting completed views are then directly passed to the pretrained DA3-Giant model~\cite{lin2026depth} to estimate the camera parameters and reconstruct 3D Gaussian scenes. All components of these baselines are kept frozen, and we use the officially recommended inference settings without tuning on the test sets.

For the comparison, all methods use the same input-view sets, masked inputs, image resolution, and pose-free evaluation protocol. Neither ground-truth nor externally estimated camera poses are provided as model inputs. These baselines instantiate two straightforward cascaded paradigms, namely ``2D inpainting + pose-free 3D reconstruction'' and ``multi-view inpainting + pose-free 3D reconstruction.'' Their purpose is to examine whether combining an off-the-shelf image inpainter with a generic 3D foundation model is sufficient for this task. In contrast, FreeInpaint directly adapts the 3D foundation model to masked observations and performs mask-aware cross-view reasoning for camera estimation, geometry recovery, and 3D scene inpainting.

\subsection{More Analysis}
\subsubsection{Discussion on Support Token Refinement}
We further analyze the design choices and parameter settings of our support token refinement strategy. Since this mechanism is specifically triggered to handle severe occlusions where the reference view lacks sufficient visual cues, we conduct the following ablation studies on six challenging scenes selected from the 360-USID~\cite{wu2025aurafusion360} and IMFine~\cite{shi2025imfine} datasets.

\noindent\textbf{Ablation on Support View Generation.}
When refinement is applied, we find that leveraging the IP-Adapter without SDEdit repairs the basic geometry of the bottle but introduces structural artifacts due to multi-view inconsistencies in the generated anchor view. In addition, using SDEdit without the IP-Adapter maintains spatial structure but generates incorrect textures that fail to match the original reference. By combining both priors, our full method successfully repairs geometric distortions while maintaining textural identity without introducing extra artifacts.

\begin{figure}[h]
    \centering
    \includegraphics[width=\linewidth]{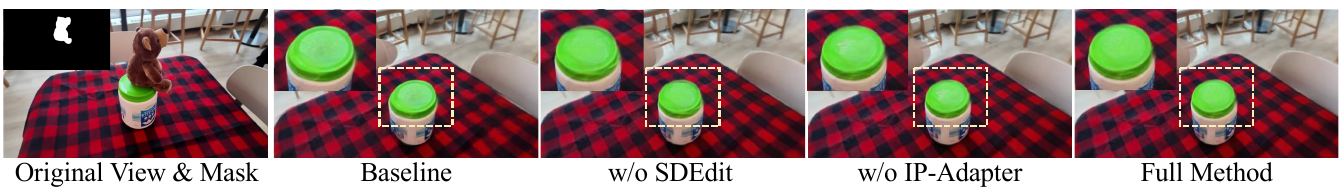}
    \caption{\small \textbf{Ablation on the Support View Generation.} Our full method effectively reconstructs the occluded regions with high fidelity, preventing the geometric or textural inconsistencies seen in the ablated baselines.}
    \label{fig:suppl_ab}
\end{figure}

\begin{table*}[h]
    \centering
    \caption{\small \textbf{Ablation studies on parameters of support token refinement.} We evaluate the impact of different iteration numbers and visibility score thresholds ($\tau$) on the 360-USID~\cite{wu2025aurafusion360} and IMFine~\cite{shi2025imfine} datasets. \textbf{Note that the reported metrics are averaged across both datasets.} ``Iters.'' refers to the total number of feed-forward passes through our model (e.g., 3 Iters. equals 1 initial pass plus up to 2 refinement loops). The default settings used in our main experiments are highlighted in \colorbox{gray!20}{gray}.}
    \label{tab:iterative_ablations}
    
    \begin{minipage}[t]{0.48\textwidth}
        \centering
        \setlength{\tabcolsep}{6pt}
        \resizebox{\linewidth}{!}{
        \begin{tabular}{c|cc|c}
            \toprule[1.2pt]
            \textbf{Max Iters.} & \textbf{PSNR} $\uparrow$  & \textbf{LPIPS} $\downarrow$ & \textbf{Time} $\downarrow$ \\
            \midrule
            $1$ (Initial pass)  &18.02   &0.2738&  0.41   \\
            $2$  &18.29   &0.2703&  1.12   \\
            \rowcolor{gray!20}
            $3$  &18.46   &0.2623 &  1.83   \\
            $4$  & 18.48  &0.2620&  2.54   \\
            \bottomrule[1.2pt]
        \end{tabular}}
        \centerline{\small (a) Impact of maximum iteration numbers.}
    \end{minipage}
    \hfill
    \begin{minipage}[t]{0.48\textwidth}
        \centering
        \setlength{\tabcolsep}{3pt}
        \resizebox{\linewidth}{!}{
        \begin{tabular}{c|cc|c}
            \toprule[1.2pt]
            \textbf{Threshold} $\tau$ & \textbf{PSNR} $\uparrow$ & \textbf{LPIPS} $\downarrow$ & \textbf{Avg. Iters} $\downarrow$ \\
            \midrule
            $0.93$  &18.12 &0.2702  &  1.33   \\
            $0.95$  &18.25 & 0.2689 &  1.50   \\
            \rowcolor{gray!20}
            $0.97$  &\textbf{18.46 }  &\textbf{0.2623} &   2.67  \\
            $0.99$  &18.30 &0.2629  &   4.33  \\
            \bottomrule[1.2pt]
        \end{tabular}}
        \centerline{\small (b) Impact of visibility score threshold.}
    \end{minipage}
\end{table*}

\noindent\textbf{Ablations on Iterative Numbers.}
As shown in Tab.~\ref{tab:iterative_ablations} (a), relying purely on a single-pass inference ($1$ iter) without support token refinement yields relatively lower metrics, as the feed-forward model inherently struggles with complete blind spots in scenes with large view variations. Increasing the maximum iterations to 2 or 3 (i.e., triggering 1 or 2 refinement loops) significantly improves PSNR and LPIPS, demonstrating the effectiveness of injecting generative support evidence for unseen geometries. However, beyond 3 iterations, the performance yields marginal gains (e.g., only $+0.02$ dB in PSNR from iter 3 to 4) while incurring a linear increase in computational overhead. Therefore, we set $3$ as the maximum iteration limit to offer the optimal trade-off between high-fidelity inpainting quality and inference efficiency.

\noindent\textbf{Ablations on Threshold of Visibility Score.}
In Tab.~\ref{tab:iterative_ablations} (b), we further analyze the sensitivity of the visibility score threshold $\tau$. A lower threshold fails to trigger the refinement for severe blind spots, leading to under-completion. In contrast, an excessively high threshold (e.g., $\tau=0.99$) forces the model to heavily rely on 2D diffusion outputs, overwriting well-reconstructed regions. This over-refinement inevitably breaks the inherent 3D consistency of the foundation model and unnecessarily inflates the average iteration count (increasing to 4.33 iters). Therefore, we empirically set $\tau=0.97$, which yields the optimal balance by accurately locating true geometric holes without disrupting reliable multi-view observations.

\begin{figure}[t]
    \centering
    \includegraphics[width=\linewidth]{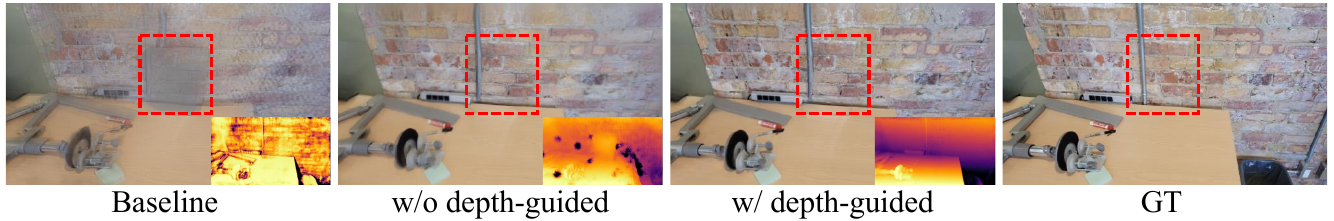}
    \vspace{-10pt}
    \caption{\small Visualization of Fine-Tuning Strategy on the SPIn-NeRF~\cite{mirzaei2023spin} dataset.}
    \label{fig:ft}
\end{figure}
\noindent\textbf{Effect of the LCM-based Diffusion Prior on Support Token Refinement.} As shown in Tab.~\ref{tab:lcm_ablation}, we evaluate the impact of the LCM-based diffusion prior on three scenes from the 360-USID dataset~\cite{wu2025aurafusion360} that require additional confidence-based support token refinement. We compare our 4-step LCM pipeline against a standard 50-step diffusion baseline. For the runtime analysis, we measure a single refinement iteration, which encompasses two forward passes of FreeInpaint. Notably, the LCM prior reduces the diffusion process time to just 0.3s, demonstrating a significant speedup over the 1.6s required by the standard baseline. Furthermore, the LCM-based approach yields better PSNR and LPIPS scores on the tested scenes. These results demonstrate the effectiveness of integrating an LCM-based prior into the support token refinement strategy.

\noindent\textbf{Discussion on Fine-Tuning Strategy.}
We further analyze the impact of depth-guided self-distillation within our fine-tuning strategy on the SPIn-NeRF~\cite{mirzaei2023spin} dataset. As illustrated in Fig.~\ref{fig:ft}, fine-tuning without depth-guided self-distillation leads to noticeable geometric artifacts, such as unnatural holes or flattened surfaces within the inpainted regions. In contrast, introducing depth-guided self-distillation effectively constrains the 3D geometry. This not only yields artifact-free geometric inpainting but also significantly improves the texture quality. Because accurate underlying geometry provides a robust foundation for optimizing the 3D Gaussians, it naturally translates to higher-fidelity texture rendering. The fine-tuning with depth-guided self-distillation demonstrates better overall visual quality compared to the baseline lacking depth constraints.

\noindent\textbf{Effect of Different Mask-Aware Attention Strategies.}
We further extend the ablation study in Tab.~\ref{tab:learnable_mask_attention} by implementing MAT-style~\cite{li2022mat} and MLLAM-style~\cite{barrios2024multi} masking strategies under the same DA3 backbone, training protocol, and Support Token Refinement, while varying only the mask-attention mechanism. The MAT-style variant uses rule-based dynamic binary mask propagation, whereas the MLLAM-style variant applies layer-wise attention modulation without initialization from the input binary mask. As shown in Tab.~\ref{tab:learnable_mask_attention_2}, both variants outperform the baseline without mask attention, while our mask-initialized, feature-conditioned learnable mask attention strategy achieves the best performance in both scene inpainting and camera pose estimation.

\begin{figure}[t]
    \centering
    \includegraphics[width=\linewidth]{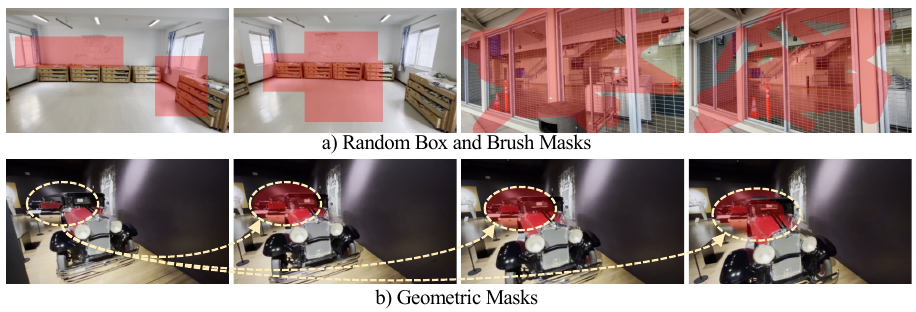}
    \vspace{-5pt}
    \caption{\small Visualization of Mask Generation Strategy.}
    \label{fig:suppl_mask}
    \vspace{-10pt}
\end{figure}

\begin{table*}[h]
    \centering
    \begin{minipage}[t]{0.48\textwidth}
        \centering
        \caption{\small \textbf{Effect of the LCM-based diffusion prior} on support view generation. ($\cdot$) denotes the time required for two forward passes of FreeInpaint.}
        \vspace{2pt}
        \label{tab:lcm_ablation}
        \resizebox{\linewidth}{!}{
        \begin{tabular}{l|ccc}
            \toprule[1.2pt]
            \textbf{Methods} & \textbf{PSNR} $\uparrow$ & \textbf{LPIPS} $\downarrow$ & \textbf{Time} $\downarrow$ \\ 
            \midrule
            w/o LCM (50 steps) & 18.22 & 0.2736 & 1.6s (+0.8s) \\
            w/ LCM (4 steps) & \textbf{18.67} & \textbf{0.2689} & \textbf{0.3s (+0.8s)} \\
            \bottomrule[1.2pt]
        \end{tabular}}
    \end{minipage}
    \hfill
    \begin{minipage}[t]{0.48\textwidth}
        \centering
        \caption{\small \textbf{Ablation studies on Mask Generation Strategy}, evaluated on the SPIn-NeRF dataset~\cite{mirzaei2023spin}.}
        \vspace{2pt}
        \label{tab:mechanism_ablations}
        \resizebox{\linewidth}{!}{
        \begin{tabular}{l|ccc}
            \toprule[1.2pt]
            \textbf{Mask Strategy} & \textbf{PSNR} $\uparrow$ & \textbf{SSIM} $\uparrow$ & \textbf{LPIPS} $\downarrow$ \\
            \midrule
            Random Box and Brush Only  & 17.18  & 0.3652  & 0.3022  \\
            Geometric Only       & 17.71  & 0.3835  & 0.2877  \\
            \rowcolor{gray!20}
            \textbf{Hybrid (Ours)}  & \textbf{17.79}  & \textbf{0.3869}  & \textbf{0.2819}  \\
            \bottomrule[1.2pt]
        \end{tabular}}
    \end{minipage}
\end{table*}

\begin{table}[!t]
    \centering
    \small
    \setlength{\tabcolsep}{15pt}
    \caption{\small Effect of different mask-aware attention strategies under the same DA3 backbone and training protocol.}
    \label{tab:learnable_mask_attention_2}
    \begin{tabular}{l|ccccc}
    \toprule
    Mask Attention & PSNR$\uparrow$ & LPIPS$\downarrow$ & ATE$\downarrow$ & RPE$_t$$\downarrow$ & RPE$_r$$\downarrow$ \\
    \midrule
    MAT-style mask~\cite{li2022mat}      & 18.13& 0.2620 & 0.0361 & 0.0743 & 0.7459 \\
    MLLAM-style mask~\cite{barrios2024multi}     & 18.06 & 0.2651 & 0.0374 & 0.0775 & 0.7468 \\ \midrule
    \textbf{Our Learnable mask}  & \textbf{18.58} & \textbf{0.2526} & \textbf{0.0340} & \textbf{0.0644} & \textbf{0.7069} \\
    \bottomrule
    \end{tabular}
\end{table}

\begin{table*}[t]
    \centering
    \small
    \setlength{\tabcolsep}{6pt}
    \caption{\small View-count scalability comparison with InstaInpaint~\cite{you2025instainpaint} on SPIn-NeRF~\cite{mirzaei2023spin} and 360-USID~\cite{wu2025aurafusion360}. ``Total Input Views'' includes one reference view and the remaining masked source views.}
    \label{tab:viewscaling_instainpaint}
    \begin{tabular}{c|l|ccc|ccc}
    \toprule
    \multirow{2}{*}{\shortstack{Total Input\\Views}} & \multirow{2}{*}{Method} & \multicolumn{3}{c|}{SPIn-NeRF}& \multicolumn{3}{c}{360-USID} \\ 
    & & PSNR$\uparrow$& LPIPS$\downarrow$& FID$\downarrow$
    & PSNR$\uparrow$& LPIPS$\downarrow$& FID$\downarrow$ \\
    \midrule

    \multirow{2}{*}{4} & InstaInpaint & 16.89 & 0.2844 & 149.11 & 14.66 & 0.4934 & 259.39 \\
    & \textbf{FreeInpaint (Ours)} & \textbf{17.79} & \textbf{0.2819} & \textbf{148.05} & \textbf{17.19} & \textbf{0.3148} & \textbf{226.30} \\
    \midrule

    \multirow{2}{*}{8} & InstaInpaint & 16.72 & 0.3013 & 144.31
    & 15.60 & 0.4707 & 228.08 \\
    & \textbf{FreeInpaint (Ours)} & \textbf{18.32} & \textbf{0.2702} & \textbf{143.88} & \textbf{18.58} & \textbf{0.2526} & \textbf{199.61} \\
    \midrule

    \multirow{2}{*}{16} & InstaInpaint & 17.04 & 0.3630 & 168.35
    & 15.51 & 0.5012 & 265.28 \\
    & \textbf{FreeInpaint (Ours)} & \textbf{18.33} & \textbf{0.2688} & \textbf{145.97} & \textbf{18.66} & \textbf{0.2512} & \textbf{196.67} \\
    \midrule

    \multirow{2}{*}{32}
    & InstaInpaint & 17.12 & 0.4186 & 239.05 & 14.58 & 0.5330 & 268.26 \\
    & \textbf{FreeInpaint (Ours)} & \textbf{18.47} & \textbf{0.2683} & \textbf{146.44} & \textbf{18.69} & \textbf{0.2519} & \textbf{194.32} \\
    \bottomrule
    \end{tabular}
\end{table*}

\begin{table}[!t]
    \centering
    \small
    \setlength{\tabcolsep}{8pt}
    \caption{\small Quantitative evaluation on unseen in-the-wild scenes from GS25~\cite{khalid20243dego} and Co3D~\cite{reizenstein2021common}.}
    \label{tab:in_the_wild}
    \begin{tabular}{l|c|cc|cc}
    \toprule
    \multirow{2}{*}{Method} & \multirow{2}{*}{Pose Input}&\multicolumn{2}{c|}{GS25~\cite{khalid20243dego}} & \multicolumn{2}{c}{Co3D~\cite{reizenstein2021common}} \\
    & & C-KID$\downarrow$ & C-FID$\downarrow$ & C-KID$\downarrow$
    & C-FID$\downarrow$ \\
    \midrule
    LaMa~\cite{suvorov2022resolution}+DA3 & \xmark& 0.3030 & 305.23 & 0.3870 & 360.72 \\
    MVInpainter~\cite{cao2024mvinpainter}+DA3& \xmark&0.3624 & 356.07& 0.3975 & 361.37 \\
    InstaInpaint~\cite{you2025instainpaint} &\cmark& 0.3406 & 330.84& 0.3568 & 330.41\\ \midrule
    \textbf{FreeInpaint (Ours)}& \xmark&\textbf{0.2889} & \textbf{292.28}& \textbf{0.3295} & \textbf{325.44} \\
    \bottomrule
    \end{tabular}
\end{table}

\subsection{Limitations and Future Work}
\begin{figure}[!t]
    \centering
    \includegraphics[width=\linewidth]{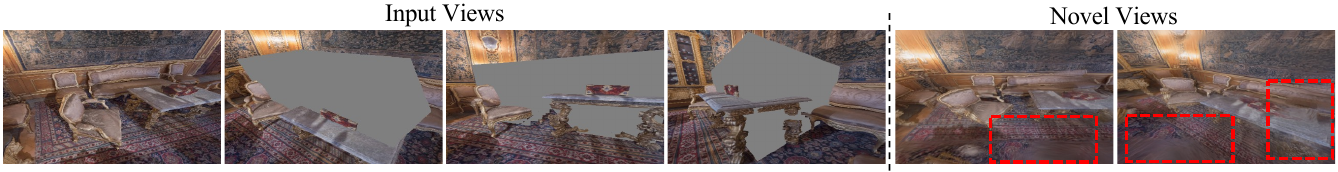}
    \caption{\small \textbf{Additional Visualization on the NeRFiller~\cite{weber2024nerfiller} dataset.} As illustrated, in extreme scenarios where highly sparse input views are combined with large-scale masks, the model struggles to retrieve sufficient contextual information. This occasionally degrades the inpainting quality and introduces some visual artifacts.}
    \label{fig:sm_failure}
    \vspace{-10pt}
\end{figure}

While FreeInpaint establishes a robust framework for pose-free 3D scene inpainting, it has some limitations. Primarily, our support token refinement relies on 2D diffusion models to inpaint blind spots, making the final 3D quality inherently bounded by the performance of the 2D prior. Future research could integrate native 3D or multi-view consistent video diffusion priors to ensure more stable geometric consistency. Furthermore, while our method excels with sparse, unposed collections, extreme cases involving only 1 or 2 restricted views of a complex 360-degree scene remain challenging (In Fig.~\ref{fig:sm_failure}) due to error accumulation during support token refinement. To address this, future iterations could incorporate a persistent spatial memory mechanism to maintain long-term geometric consistency across arbitrary refinement steps. In addition, support token refinement could be further extended to sparse-view reconstruction and novel view synthesis, where insufficiently observed regions could be identified and refined using video diffusion priors.

\begin{figure}[!t]
    \centering
    \includegraphics[width=\linewidth]{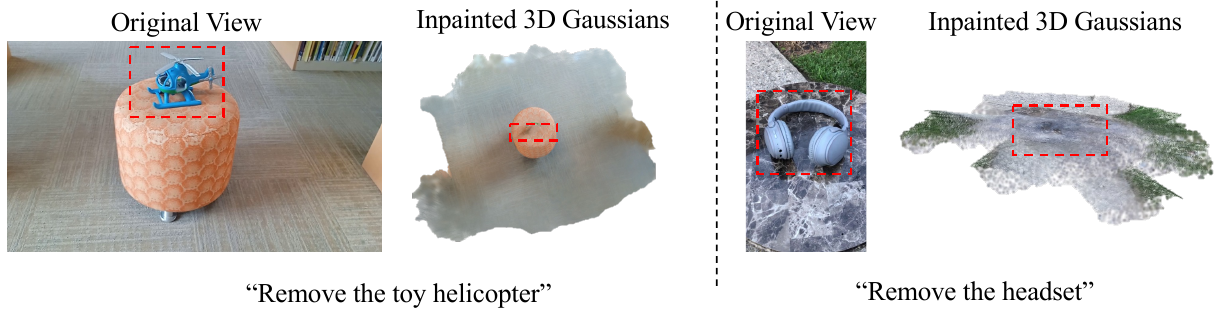}
    \caption{\small Additional Qualitative Results on the GS25 dataset~\cite{khalid20243dego} under unposed 8-view in-the-wild settings.}
    \label{fig:sm_inthewild}
    \vspace{-15pt}
\end{figure}

\begin{figure}[t]
    \centering
    \includegraphics[width=\linewidth]{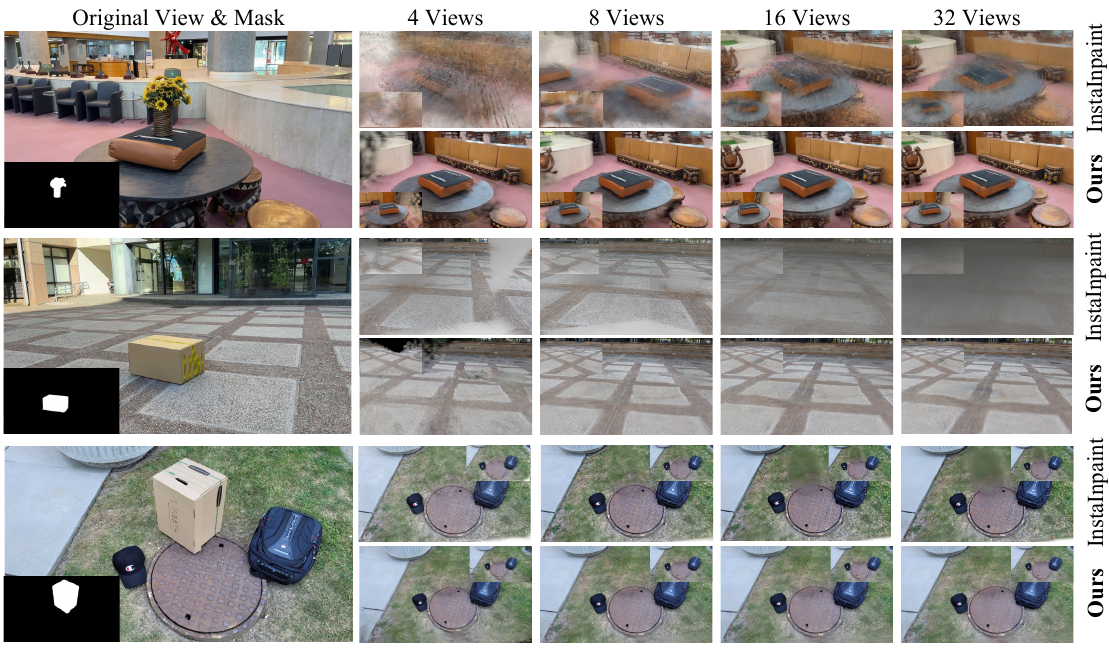}
    \caption{\small Additional qualitative results on view-count scalability.}
    \label{fig:sm_view_scaling}
\end{figure}

\begin{figure}[t]
    \centering
    \includegraphics[width=\linewidth]{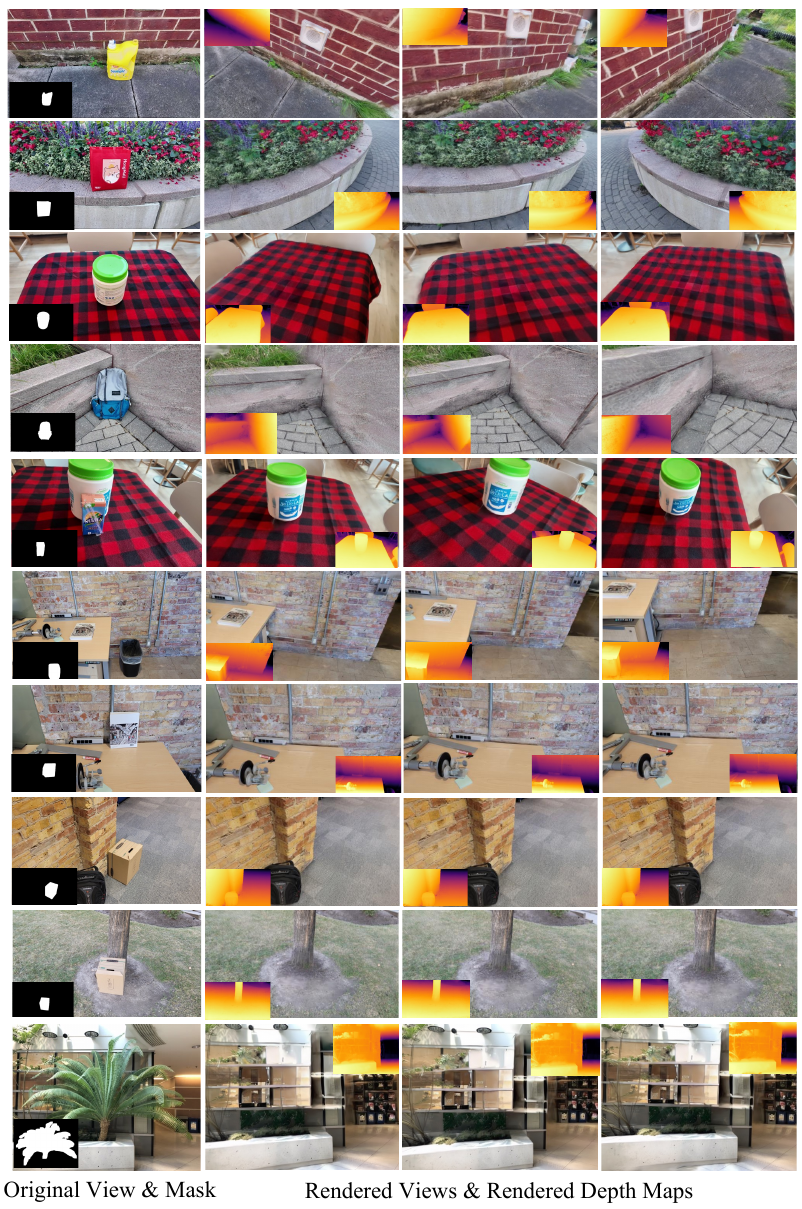}
    \caption{\small Additional Qualitative Results on the SPIn-NeRF~\cite{mirzaei2023spin}, LLFF~\cite{mildenhall2019llff}, and IMFine~\cite{shi2025imfine} datasets.}
    \label{fig:sm_vis}
\end{figure}

\begin{figure}[t]
    \centering
    \includegraphics[width=\linewidth]{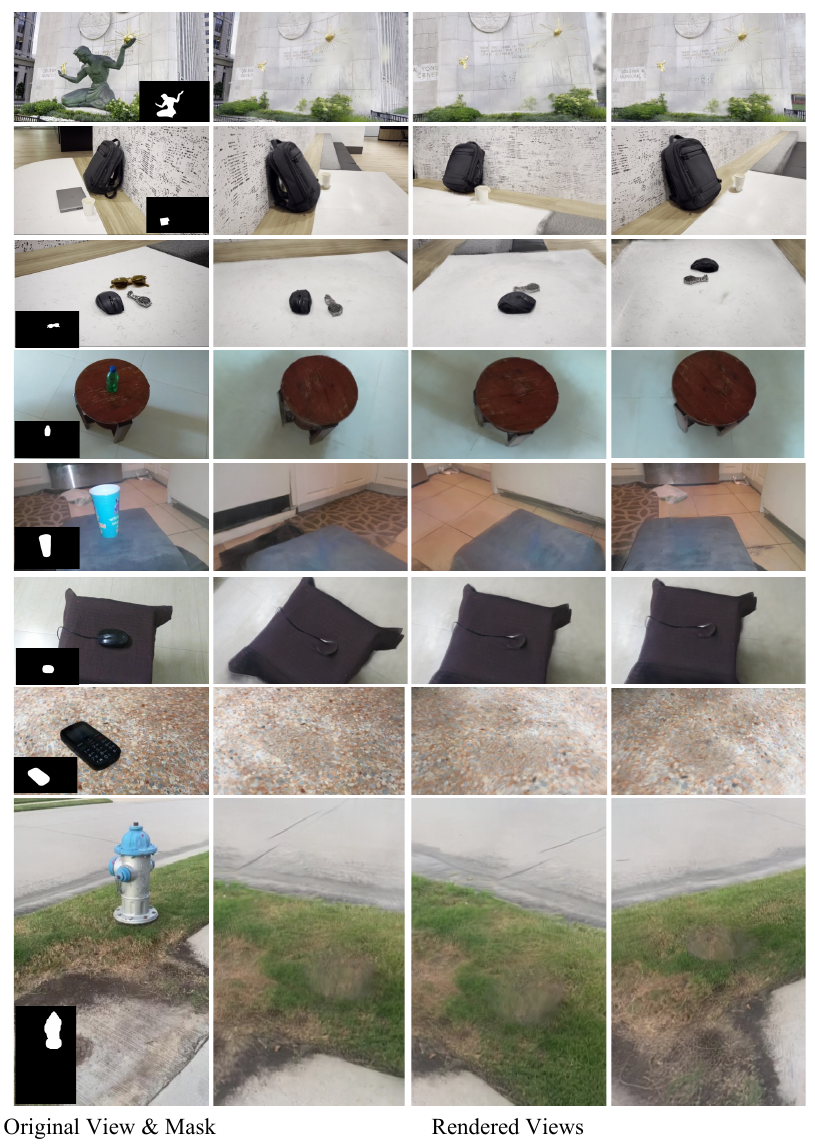}
    \caption{\small Additional Qualitative Results on GS25~\cite{khalid20243dego} and Co3D~\cite{reizenstein2021common} datasets.}
    \label{fig:sm_vis_inthewild}
\end{figure}

\subsection{More Results}
\noindent\textbf{Additional Comparisons.}
We additionally compare FreeInpaint with InstaInpaint~\cite{you2025instainpaint}, a feed-forward 3D scene inpainting method that requires camera poses as input. As shown in Tab.~\ref{tab:viewscaling_instainpaint}, FreeInpaint consistently outperforms InstaInpaint across both datasets and all evaluated metrics. As further illustrated in Fig.~\ref{fig:sm_view_scaling}, FreeInpaint benefits from additional input views, with reconstruction quality improving while perceptual quality remains stable. In contrast, InstaInpaint does not consistently benefit from denser inputs, with LPIPS and FID degrading notably at 16 and 32 views. These results demonstrate that FreeInpaint can more robustly exploit denser collections of unposed observations without requiring camera poses as model inputs.

To further evaluate generalization to in-the-wild scenes, we compare LaMa~\cite{suvorov2022resolution}+DA3, MVInpainter~\cite{cao2024mvinpainter}+DA3, InstaInpaint~\cite{you2025instainpaint}, and FreeInpaint on 10 GS25~\cite{khalid20243dego} scenes and 15 Co3D~\cite{reizenstein2021common} scenes. For each scene, all methods use the same eight input views, consisting of one diffusion-inpainted reference image and seven masked source views. Since these datasets do not provide ground-truth backgrounds after object removal, we adopt the same no-reference evaluation protocol used for LLFF and compute C-KID and C-FID within the object-mask bounding boxes across all evaluation views. As shown in Tab.~\ref{tab:in_the_wild}, FreeInpaint achieves the best results across all four metrics, providing further evidence of its generalization to unseen casual captures.

\textbf{More Visual Results.}
In this section, we present additional visual results on the SPIn-NeRF~\cite{mirzaei2023spin}, LLFF~\cite{mildenhall2019llff}, and IMFine~\cite{shi2025imfine} datasets. As illustrated in Fig.~\ref{fig:sm_vis}, we show rendered views and rendered depth maps across diverse scenes, demonstrating that our method achieves multi-view consistent texture and geometry. We further evaluate our method on the in-the-wild GS25 dataset~\cite{khalid20243dego} and Co3D~\cite{reizenstein2021common} dataset under the unposed setting with eight input views. As shown in Figs.~\ref{fig:sm_inthewild} and \ref{fig:sm_vis_inthewild}, FreeInpaint remains robust across diverse in-the-wild scenes.

\subsection{Impact Statement}
Our work focuses on the field of 3D generative models with important ethical implications, including the risk of generating misleading or manipulated visual content. We recognize these challenges and affirm our commitment to responsible research and development. While this technology holds promise for empowering creative professionals and democratizing 3D content creation, we are dedicated to proactively addressing ethical concerns in its deployment.

\clearpage


\end{document}